\documentclass[a4paper,twoside]{article}

\usepackage{epsfig}
\usepackage{subcaption}
\usepackage{calc}
\usepackage{amssymb}
\usepackage{amsbsy}
\usepackage{amstext}
\usepackage{amsmath}
\usepackage{amsthm}
\usepackage{multicol}
\usepackage{pslatex}
\usepackage{apalike}
\usepackage{algorithm2e}
\usepackage[bottom]{footmisc}
\usepackage{SCITEPRESS}     % Please add other packages that you may need BEFORE the SCITEPRESS.sty package.
\usepackage{graphicx}
\begin{document}

\title{Unscented Transform-based Pure Pursuit Path-Tracking Algorithm under Uncertainty}

\author{\authorname{Chinnawut Nantabut\sup{1}\orcidAuthor{0000-0002-5767-6023}}
\affiliation{\sup{1}The
Sirindhorn International Thai-German Graduate School of Engineering, King Mongkut's University of Technology North Bangkok, 1518 Pracharat 1 Road, Wongsawang, Bangsue, Bangkok, Thailand}
\email{chinnawut.n@tggs.kmutnb.ac.th}
}


\abstract{Automated driving has become more and more popular due to its potential to
eliminate road accidents by taking over driving tasks from humans.
One of the remaining challenges is to follow a planned path autonomously,
especially when uncertainties in self-localizing or understanding the surroundings
can influence the decisions made by autonomous vehicles,
such as calculating how much they need to steer to minimize tracking errors.
In this paper, a modified geometric pure pursuit path-tracking algorithm
is proposed, taking into consideration such uncertainties using the unscented transform.
The algorithm is tested through simulations for typical road geometries, 
such as straight and circular lines.}

\onecolumn \maketitle \normalsize \setcounter{footnote}{0} \vfill

\section{\uppercase{Introduction}}
\label{sec:introduction}

Significant research in the field of autonomous vehicles seeks to reduce injuries and fatalities by fully 
replacing human decision-making in driving. Despite the recent surge in the popularity of self-driving cars, 
research into tracking algorithms - where vehicles attempt to stay close to a desired reference path - has been ongoing for decades.

Several approaches, including geometry-based, robust, optimization-based, and AI-based algorithms, 
are commonly employed to address this challenge \cite{Rokonuzzaman21}. 
Geometry-based algorithms, such as pure pursuit controllers, remain popular due to their simplicity. 
These algorithms rely solely on geometric information from the vehicle and its current position relative to the tracked path, 
allowing timely decisions to handle potential hazards. Recent tests have applied this algorithm in driver assistance systems 
\cite{Wang17} and agricultural field tests \cite{Qiang21}. 

However, geometry-based algorithms are most effective when vehicle limitations are not exceeded, 
such as during highway driving or aggressive steering maneuvers, as demonstrated by \cite{Nan21}. 
In scenarios where vehicle constraints are not fulfilled, optimization-based algorithms 
like Linear Quadratic Regulator (LQR), Model Predictive Control (MPC), 
and AI-based methods - such as neural networks that can model vehicle behavior 
more precisely - are often better suited. In typical cases, such as low-speed urban traffic, 
a kinematic bicycle model combined with geometric-based algorithms is sufficient. For instance, \cite{Lee23} showed that pure pursuit 
algorithms can achieve similar tracking performance to LQR and MPC in low-friction conditions.

Though considered a traditional method, the pure pursuit algorithm continues to attract research interest. 
Much of the work explores optimal ways to determine the look-ahead distance, 
measured from the vehicle’s rear axle to the reference path.
Recent studies have also incorporated techniques from other tracking algorithms into pure pursuit. 
For example, \cite{Li23} integrated adaptive pure pursuit to enhance path planning in cluttered environments. 
These adaptations incorporate principles from more advanced tracking algorithms, helping to optimize the look-ahead distance.

In AI-based methods, \cite{Joglekar22} applied deep reinforcement learning (DRL) to skid-steered vehicles, 
enhancing adaptation in nonlinear off-road scenarios compared to MPC. 

Regarding optimization techniques, 
evolutionary algorithms are often used. For example, \cite{Wang20} employed the Salp Swarm Algorithm (SSA) 
to optimize the look-ahead distance for improved tracking, while \cite{Wu21} 
used Particle Swarm Optimization (PSO) to adjust preview distances based on vehicle speed and track curvature.
Similarly, \cite{Yang22} assessed lateral and heading errors for steering optimization, 
while \cite{Zhao24} applied a biomimetics-inspired adaptive pure pursuit for agricultural machinery. 

Control theory approaches are also used; for instance, \cite{Huang20} and \cite{Cao24} 
utilized PID controllers to optimize steering at low speeds and for dead reckoning in navigation, respectively. 
Additionally, \cite{Chen18} combined PID controllers with low-pass filters to stabilize steering during sudden changes, 
and \cite{Qiang21} employed fuzzy controllers to calculate the look-ahead distance based on tractor speed. 
Finally, \cite{Kim24} adapted the look-ahead distance using MPC to adjust for path deviation.

\begin{figure}[!h]
  \centering
  \includegraphics[scale=0.23]{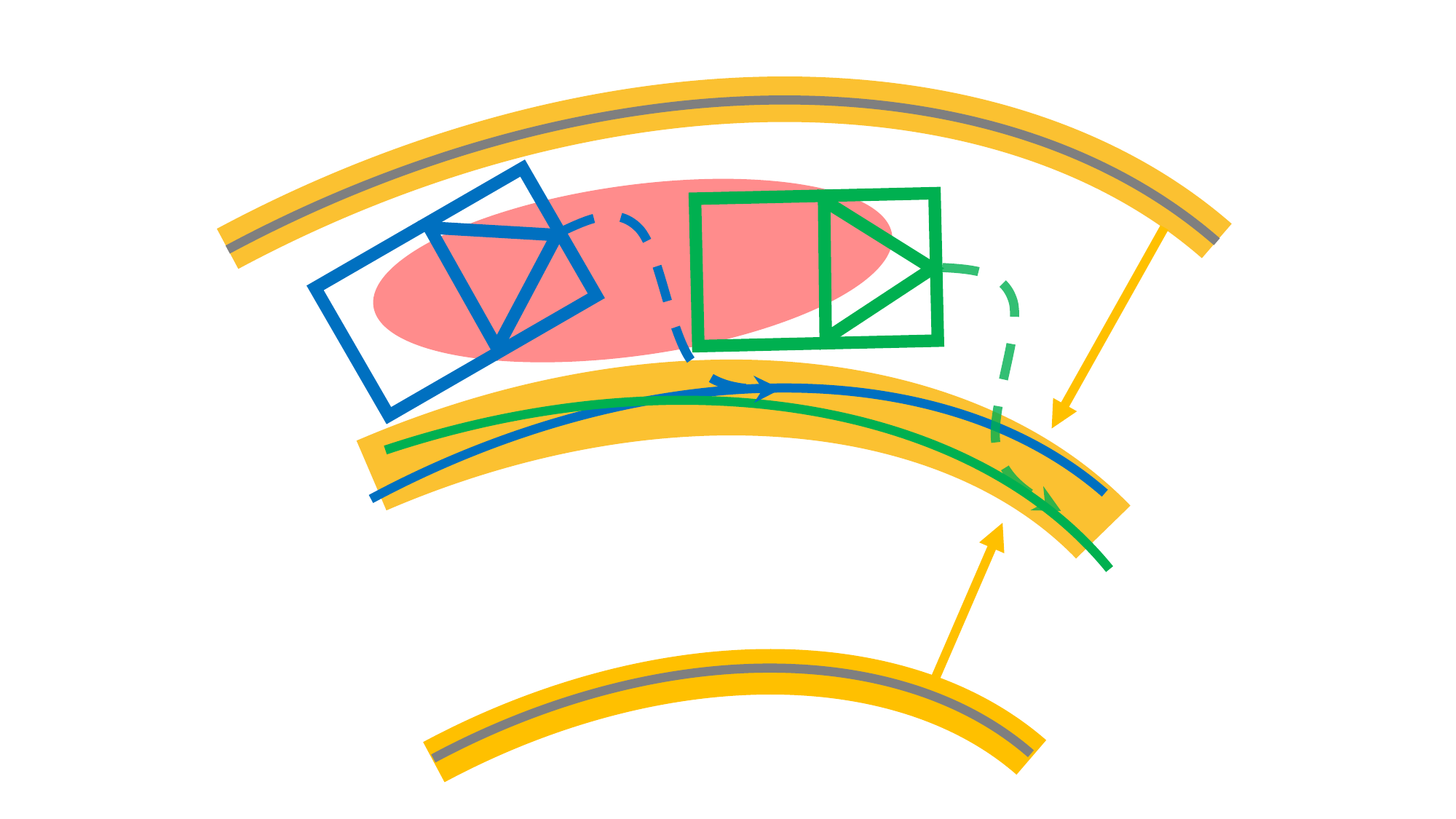}
  \caption{In the path-tracking problem, two types of uncertainty can arise:
  1) Localization uncertainty: Represented by a red confidence ellipse, this uncertainty leads 
  to variations in the vehicle’s pose (position and orientation), 
  depicted by dark blue and green lines.
  2) Detection uncertainty: Shown in yellow, 
  this arises from inaccurate identification of lane boundaries, 
  increasing uncertainty in calculating the reference path, 
  illustrated by green and blue lines.}
  \label{fig01}
\end{figure}

One remaining question is how the pure pursuit algorithm performs under uncertainty, 
as illustrated in Figure \ref{fig01}. This uncertainty can arise from GNSS inaccuracies, 
resulting in various vehicle poses (positions and orientations). 
Additionally, the accuracy of determining the reference path from 
road boundaries can be influenced by the sensing capabilities of cameras.

To date, the study by \cite{Kim23} is the only one that addresses pure pursuit 
under localization uncertainty. In their approach, they established a 95-percent 
confidence interval around the reference path and computed 
two possible steering angles based on the boundaries of this interval. 
However, relying on just two options may not be sufficient to calculate 
the optimal steering angle in the presence of uncertainty.

This work, therefore, proposes a method to derive more representative steering angles.
Given that localization noise is often assumed to follow a Gaussian distribution, 
a more robust approach for determining them - 
known as sigma points - can be derived using the unscented transform, 
a method typically applied in unscented Kalman filters \cite{Wan00}. 
These sigma points are nonlinearly transformed based on the vehicle’s geometry 
and reference path, and weighted to yield the optimal steering angle.

Furthermore, the focus is on exploring pure pursuit algorithms 
under uncertainty for common road types, 
such as straight and circular lanes \cite{Kiencke00}, 
with the goal of deriving closed-form solutions for each road type - something 
that has not been previously addressed for pure pursuit algorithms.
Clothoid lanes, which connect both straight and curved lanes, 
are excluded from this work since they can be simplified into circular cases. 
This reduction is done by first defining the waypoints 
using the method proposed by \cite{Mendez16}. 
A K-D tree is then used to identify the waypoint closest to the vehicle, 
approximately corresponding to the look-ahead distance. 
From this waypoint and its neighbors, the Menger curvature is calculated, 
and its reciprocal gives the radius of the osculating circle. 
The process for circular lanes can then be applied, as described in the subsequent section.

Additionally, it is assumed that the vehicle is aware of the type of road it is on. 
The unscented transform is employed to calculate a reasonably weighted steering 
angle for the vehicle. The proposed algorithm will be implemented in a Python 
environment to assess its performance and compare its effectiveness against 
the original pure pursuit algorithm under uncertainty across different road geometries.

\begin{figure}[!h]
   \flushleft
   \includegraphics[scale=0.22]{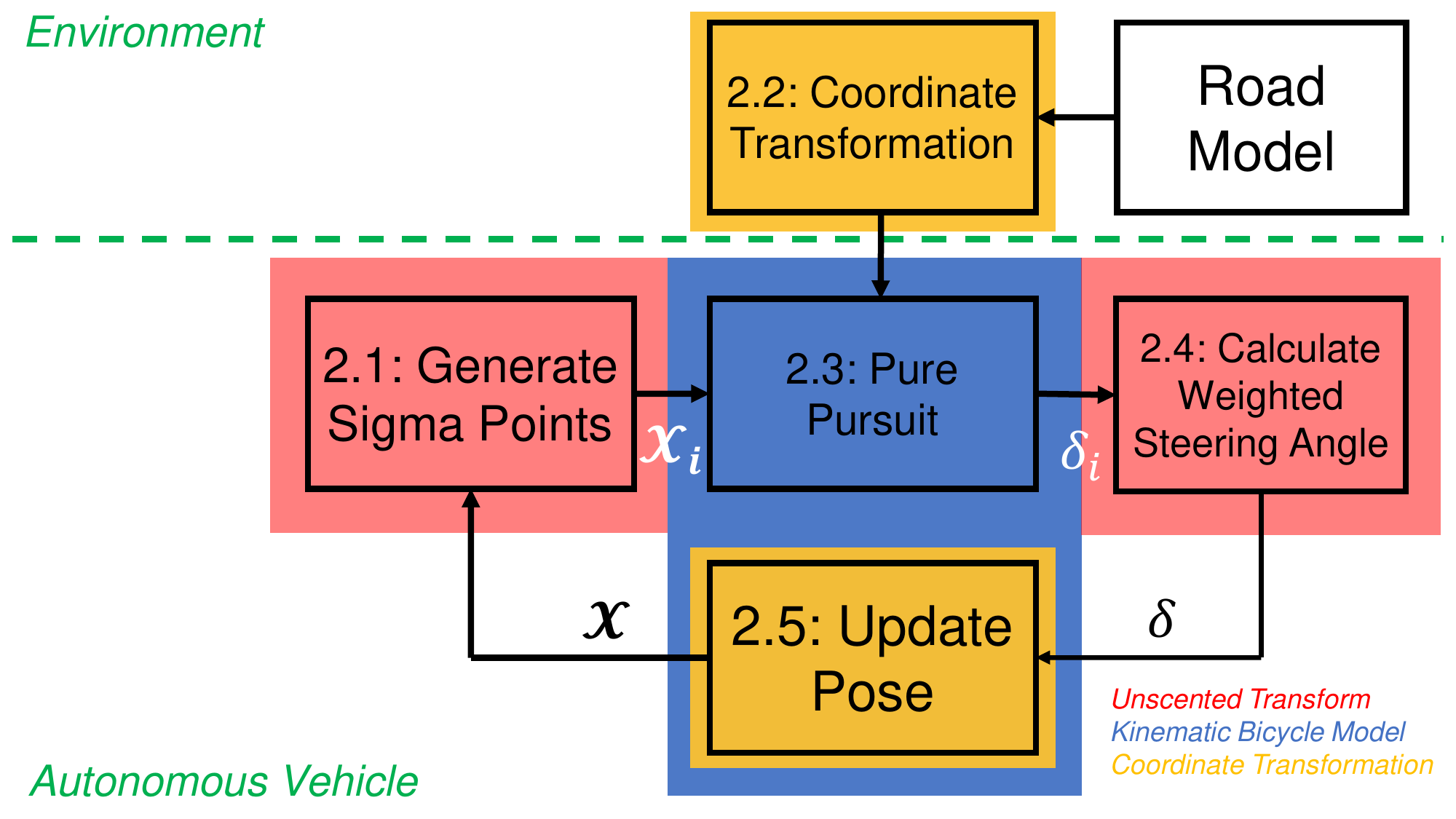}
   \caption{The general workflow of the proposed unscented transform-based 
   pure pursuit algorithm under uncertainty is as follows.}
   \label{fig02}
\end{figure}

\section{METHOD}
In Figure \ref{fig02}, the interaction between the two systems - the environment and the autonomous vehicle - 
is illustrated. The process begins with constructing the environment based on a road model, 
which consists of straight or circular lines. 
The vehicle's pose, denoted as $\pmb{\mathcal{X}}$, is then defined, and a set of sigma points, 
$\pmb{\mathcal{X}_i}$, is generated, as described in Section \ref{sectionA}. 
Subsequently, the road model is transformed from global coordinates to the vehicle’s coordinate frame, 
as explained in Section \ref{sectionB}.

These sigma points, along with the transformed road, 
are then used to calculate the new steering angles, $\delta_i$, 
using a pure pursuit algorithm based on the kinematic bicycle model, 
as detailed in Section \ref{sectionC}. The weighted steering angle is 
then computed using the unscented transform, outlined in Section \ref{sectionD}. 
Following this, the vehicle’s pose is updated using the coordinate transformation 
and the bicycle model (Section \ref{sectionE}), and the process of generating sigma points is repeated.

In summary, Sections \ref{sectionA} and \ref{sectionD} 
incorporate the unscented transform (highlighted in red), 
Sections \ref{sectionB} and \ref{sectionE} involve 
the use of coordinate transformations (highlighted in orange), 
and Sections \ref{sectionC} and \ref{sectionE} apply the kinematic bicycle model.

\subsection{Unscented Transform: Generate Sigma Points}
\label{sectionA}

As depicted in Figure \ref{fig01}, 
there are two sources of uncertainty in this problem: 
the vehicle's localization ($V$) and the computation of the reference path ($R$).

\begin{figure}[!h]
   \flushleft
   \includegraphics[scale=0.23]{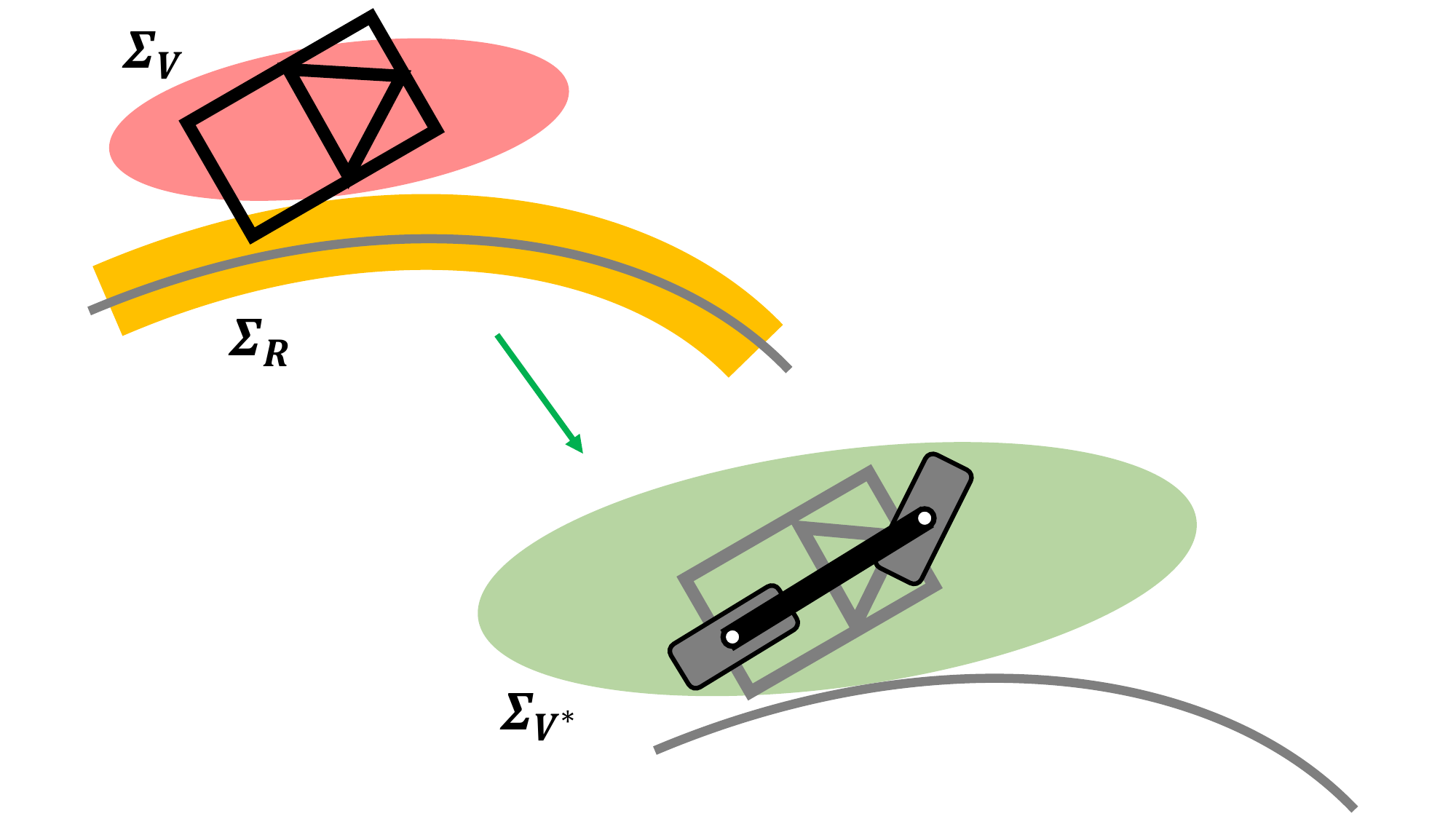}
   \caption{The calculation under uncertainty is simplified by consolidating 
   the various types of uncertainty—specifically, the vehicle's uncertainty 
   ($\pmb{\Sigma_V}$) and the road's uncertainty ($\pmb{\Sigma_R}$) - 
   into a single representation centered around the vehicle, denoted as $\pmb{\Sigma_{V^*}}$.}
   \label{fig03}
\end{figure}

As shown in Figure \ref{fig03}, the uncertainties are represented by their corresponding covariance matrices. 
The covariance matrix for the vehicle's localization is given by:
\begin{equation}
   \begin{aligned} \label{eqt99}
      \pmb{\Sigma_V} = diag(\sigma_{x,V}^2, \sigma_{y,V}^2, \sigma_{\psi,V}^2)
   \end{aligned} 
\end{equation}
and is highlighted in red. 
Conversely, the covariance matrix for the calculation 
of the reference path is represented as:
\begin{equation}
   \begin{aligned} \label{eqt99}
      \pmb{\Sigma_R} = diag(\sigma_{x,R}^2, \sigma_{y,R}^2, \sigma_{\psi,R}^2),
   \end{aligned} 
\end{equation} 
which is highlighted in yellow.
Note that $diag$ denotes a matrix with the provided values 
as its diagonal entries, with zeros elsewhere. 
Additionally, $\sigma_x$, $\sigma_y$, and $\sigma_{\psi}$ 
represent the standard deviations of the vehicle 
in the coordinates $x$, $y$, and $\psi$ respectively, 
assuming independence between them. 

As derived in \cite{Zhu19}, if these uncertainties are independent, we can sum the two covariance matrices, 
resulting in the total uncertainty being attributed solely to the vehicle's localization. 
This resulting covariance matrix is represented as:
\begin{equation}
   \begin{aligned} \label{eqt99}
      \pmb{\Sigma_{V^*}} = diag(\sigma_x^2, \sigma_y^2, \sigma_{\psi}^2) \\
      = \pmb{\Sigma_V} + \pmb{\Sigma_R},
   \end{aligned} 
\end{equation}
which is illustrated in green. For computational simplification, 
the complex vehicle is replaced by the kinematic bicycle model, 
as depicted in gray in Figure \ref{fig03}.

\begin{figure}[!h]
   \flushleft
   \includegraphics[scale=0.22]{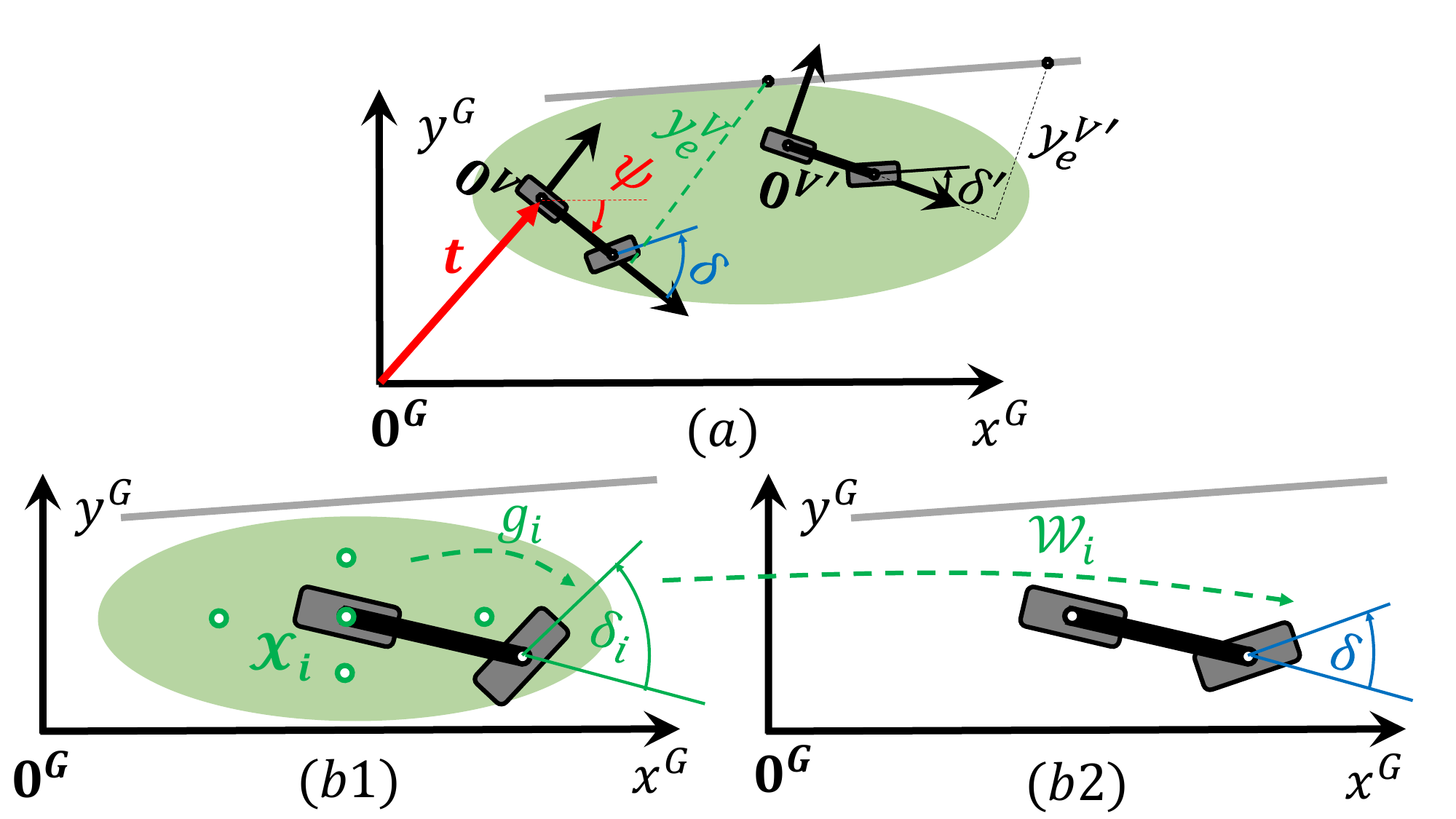}
   \caption{Tracking algorithm under uncertainty: 
   (a) Different poses result in different tracking errors, denoted as $y_e^{V}$ and $y_e^{V'}$, 
   along with corresponding steering angles, $\delta$ and $\delta'$. 
   (b) Unscented Transform: 
   (b1) Different sigma points $\pmb{\mathcal{X}_i}$ undergo a nonlinear transformation $g_i$, 
   resulting in steering angles $\delta_i$.
   (b2) These angles are then weighted by $\mathcal{W}_i$ to produce the weighted steering angle $\delta$.
}
   \label{fig04}
\end{figure}

With the incorporation of new uncertainty, 
the pure pursuit algorithm is illustrated in Figure \ref{fig04}.
The vehicle's coordinates are fixed at its rear axle, denoted by $\pmb{0^V}(x^V,y^V)$, 
and undergo translation by $\pmb{t} = [x_t \ y_t]^T$ and rotation by $\psi$ 
with respect to the global coordinates $\pmb{0^G}(x^G, y^G)$.
Due to the uncertainty $\pmb{\Sigma_{V^*}}$, the vehicle can assume multiple configurations,  
resulting in multiple possible steering angles 
$\delta$ and $\delta'$, as well as cross-track errors $y_e^V$ and $y_e^{V'}$ 
concerning the reference path in gray, as depicted in \ref{fig04} (a). 
Choosing all possible poses may be impractical.  
Therefore, according to \cite{Wan00}, 
$2L_{UT}+1$ sigma points $\pmb{\mathcal{X}_i}$
are selected from the distribution $\pmb{\Sigma_{V^*}}$, as shown in Figure \ref{fig04} (b1).
Note that $L_{UT}$ represents the 
dimensionality of the pose, which is $3$ in this case. 
Let $\pmb{\mathcal{X}} = [\pmb{t}^T \ \psi]^T$ 
represent the current pose, which serves as the "first" sigma point:
\begin{equation}
   \begin{aligned} \label{eqtX0}
      \pmb{\mathcal{X}_{0}} = \pmb{\mathcal{X}}.
   \end{aligned} 
\end{equation}

The other $2L_{UT} = 6$ sigma points are defined as:
\begin{equation}
   \begin{aligned} \label{eqtX}
      \pmb{\mathcal{X}_{1...6}} = \pmb{\mathcal{X}_0} + \sqrt{L_{UT}+\lambda_{UT}} \cdot \Delta \pmb{\mathcal{X}_j},
   \end{aligned} 
\end{equation}
where $\Delta \pmb{\mathcal{X}_{1,2}} = \pm [\sigma_x, 0, 0]^T$, 
$\Delta \pmb{\mathcal{X}_{3,4}} = \pm [0, \sigma_y, 0]^T$,
$\Delta \pmb{\mathcal{X}_{5,6}} = \pm [0, 0, \sigma_{\psi}]^T$,
and $L_{UT}$ as well as $\lambda_{UT}$ are the unscented parameters.  

These inputs, namely $\pmb{\mathcal{X}_{0}}$ and $\pmb{\mathcal{X}_{1...6}}$, 
will be utilized in the coordinate transformation and  
unscented transform steps, employing the bicycle model and pure pursuit algorithm.  
Their propose is to calculate the steering angles $\delta_i$, formulated as $g_i(\pmb{\mathcal{X}_i})$.
Subsequently, these steering angles will be weighted to form the steering angle $\delta$, 
as depicted in Figure \ref{fig04} (b2).

\subsection{Coordinate Transformation}
\label{sectionB}

Since roads are typically modeled in global coordinates, while 
the pure pursuit algorithm uses the vehicle's coordinates 
to determine the reference point using the look-ahead distance, 
the concept of coordinate transformation is introduced. 
This involves deriving equations for converting points from
global to vehicle coordinates, and vice versa.
These equations will then be applied separately for straight and circular roads.
\subsubsection{Points}
\begin{figure}[!h]
   \flushleft
   \includegraphics[scale=0.23]{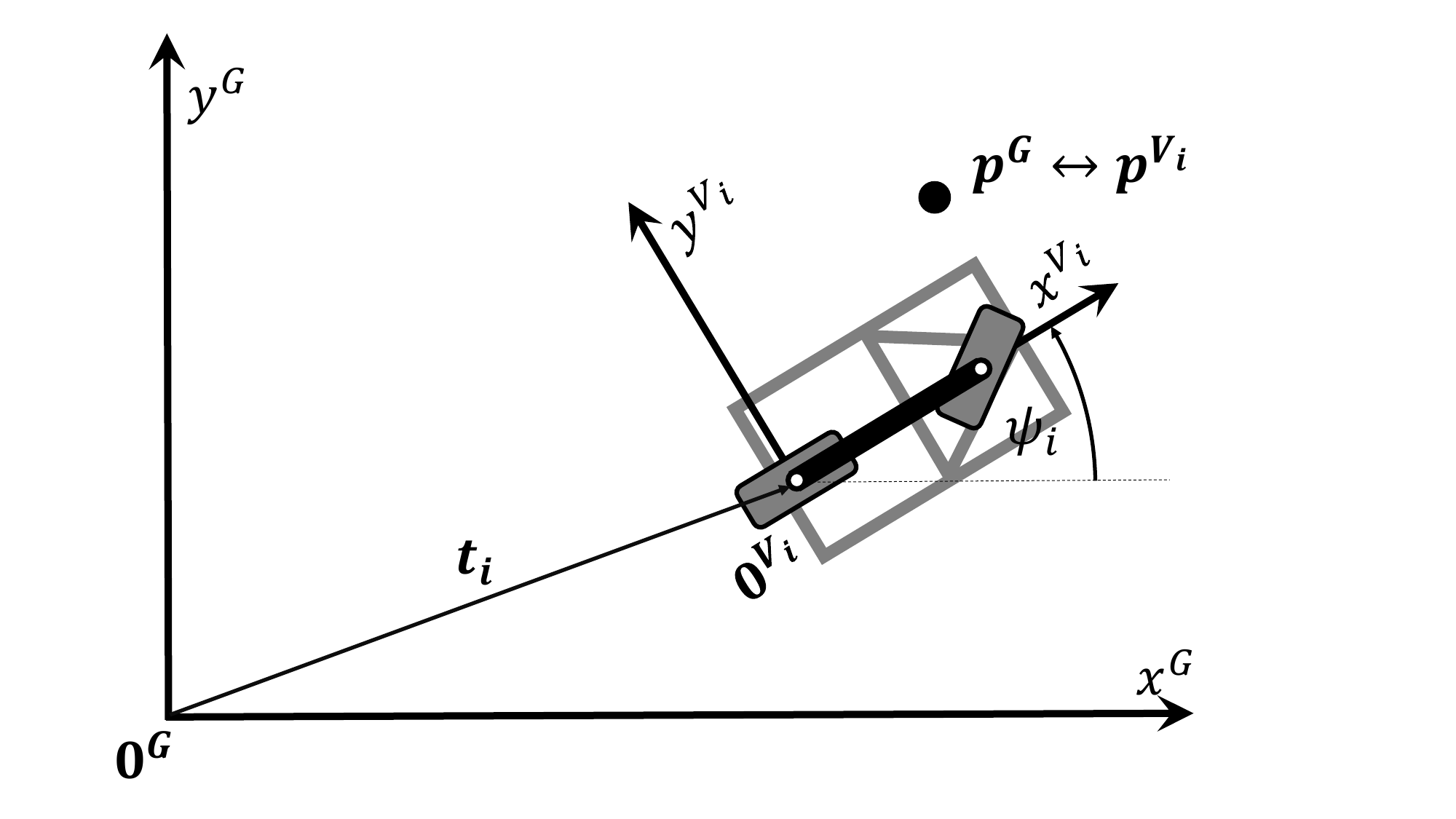}
   \caption{Transforming a point from the global coordinates, denoted by $\pmb{p^G}$, 
   into the vehicle's coordinates, denoted by $\pmb{p^V}$ and vice versa.}
   \label{fig1}
   \end{figure}
In Figure \ref{fig1}, an autonomous vehicle is located 
at position $\pmb{t_i} = [x_{t,i} \ y_{t,i}]^T$ and oriented at $\psi_i$ 
with respect to the global coordinate system $\pmb{0^G}$. This pose can correspond to  
one of the sigma points $\pmb{\mathcal{X}_i} = [\pmb{t_i}^T \ \psi_i]^T$. 
Let $\pmb{p^G}(x_p^G,y_p^G)$ and $\pmb{p^{V_i}}(x_p^{V_i},y_p^{V_i})$ represent a point $\pmb{p}$ 
described in the global 
and the $i$-th vehicle coordinates, respectively. 
We can then transform $\pmb{p^{V_i}}$ to $\pmb{p^G}$ using the following relationship: 
\begin{equation}
   \begin{aligned}
     \label{eqt3}
     \begin{bmatrix}
      x_p^G \\
      y_p^G \\
     \end{bmatrix}
     =
     \begin{bmatrix}
      \cos(\psi_i) & -\sin(\psi_i) \\
      \sin(\psi_i) & \cos(\psi_i) \\  
     \end{bmatrix}
     \begin{bmatrix}
      x_p^{V_i} \\
      y_p^{V_i} \\
     \end{bmatrix}
     +
     \begin{bmatrix}
      x_{t,i} \\
      y_{t,i} \\
     \end{bmatrix},
   \end{aligned} 
\end{equation}
or $\pmb{p^G}$ to  $\pmb{p^{V_i}}$ using the following equation:
\begin{equation}
   \begin{aligned}
     \label{eqt7}
     \begin{bmatrix}
      x_p^{V_i} \\
      y_p^{V_i} \\
     \end{bmatrix}
     =
     \begin{bmatrix}
      \cos(\psi_i) & \sin(\psi_i)  \\
      -\sin(\psi_i) & \cos(\psi_i) \\
     \end{bmatrix}
     \begin{bmatrix}
      x_p^G-x_{t,i} \\
      y_p^G-y_{t,i} \\
     \end{bmatrix}.
   \end{aligned} 
\end{equation}

Utilizing these equations, we can represent 
the roads and the sigma points in both coordinates, 
as will be discussed in the subsequent sections. 

\subsubsection{Straight Lines}
\begin{figure}[!h]
\flushleft
\includegraphics[scale=0.23]{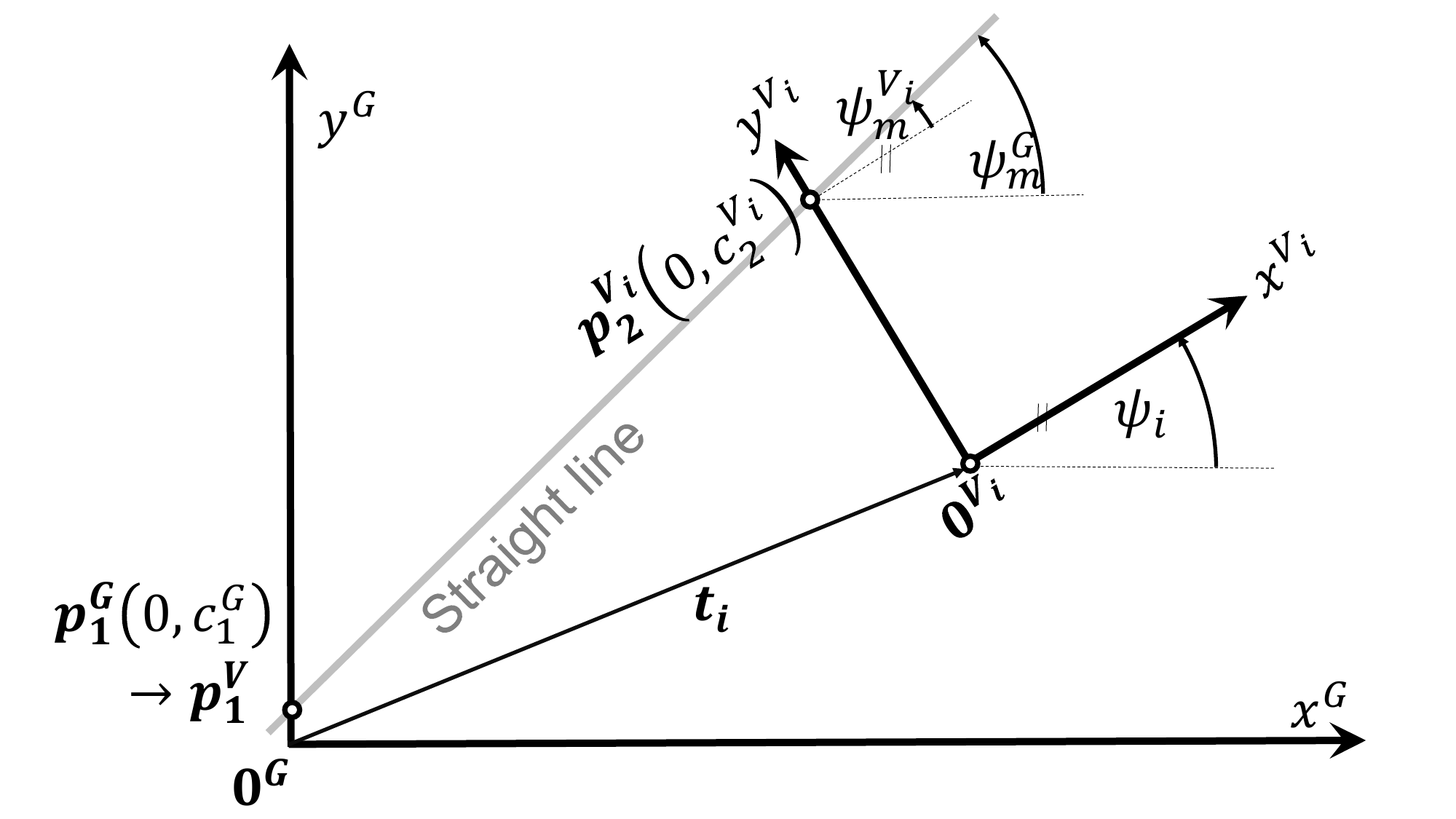}
\caption{Transforming a straight line from the global coordinates
into the vehicle's coordinates and vice versa.}
\label{fig2}
\end{figure}
Given a straight line depicted in gray, as shown in Figure \ref{fig2}, 
it can be described in the global coordinate system by:
\begin{equation}
   \begin{aligned}
     \label{eqt8}
     y^G = m^Gx^G + c_1^G,
   \end{aligned} 
 \end{equation}
 where $m^G$ denotes the slope of the straight line in the global coordinates, 
 while $c_1^G$ represents the $y^G$ intercept, indicating where the line intersects the $y^G$ axis. 
The subscript "1" states that it is located at a specific point, such as point $\pmb{p_1}$.

Our task is to transform this straight line into the $i$-th vehicle's coordinate system described by:
 \begin{equation}
   \begin{aligned}
     \label{eqt9}
     y^{V_i} = m^{V_i}x^{V_i} + c_2^{V_i}.
   \end{aligned} 
 \end{equation}

From Figure \ref{fig2}, 
 it is evident that the intercept in the $y^{V_i}$ coordinate, $c_2^{V_i}$, 
corresponds to a different point, denoted by $\pmb{p_2}$, than point $\pmb{p_1}$. 
The slope $m^{V_i}$ also needs to be calculated. 
It is worth noting that all ideal or noise-free waypoints 
must satisfy the line equation \eqref{eqt8}. 
 
 In the global coordinates, it is evident that:
 \begin{equation}
   \begin{aligned}
     \label{eqt10}
     \psi_m^G = \arctan(m^G),
   \end{aligned} 
 \end{equation}
 where $\psi_m^G$ 
 denotes the orientation of the straight line from the $x^G$ axis. 
Upon further investigation into the vehicle's coordinates, 
 the slope $m^{V_i}$ can be determined analogously using the angle $\psi_m^G$ 
 and the yaw angle $\psi_i$ as:
\begin{equation}
   \begin{aligned}
     \label{eqt11}
     m^{V_i} = \tan(\psi_m^{V_i}) = \tan(\psi_m^G-\psi_i).
   \end{aligned} 
 \end{equation}

 By utilizing \eqref{eqt7}, 
 we can transform the point $\pmb{p_1^G}(0,c_1^G)$ into the $i$-th vehicle's coordinates $\pmb{p_1^{V_i}}$ as follows:
 \begin{equation}
   \begin{aligned}
     \label{eqt12}
     \pmb{p_1^{V_i}} =  
     \begin{bmatrix}
      x_1^{V_i} \\
      y_1^{V_i} \\
     \end{bmatrix}
     =
     \begin{bmatrix}
     - x_t\cos(\psi_i) + (c_1^G - y_{t,i})\sin(\psi_i) \\
     x_t\sin(\psi_i) + (c_1^G - y_{t,i})\cos(\psi_i)  \\
     \end{bmatrix}.
   \end{aligned} 
\end{equation}

Inserting this equation into the reformulated \eqref{eqt9} yields: 
\begin{equation}
   \begin{aligned}
     \label{eqt14}
     c_2^{V_i} = y_1^{V_i}-m^{V_i}x_1^{V_i} \\
     =
     x_{t,i}(\sin(\psi_i) + m^{V_i}\cos(\psi_i)) \\ 
     + (c_1^G - y_{t,i})(\cos(\psi_i) - m^{V_i}\sin(\psi_i)).
   \end{aligned} 
\end{equation}
Using the sum formulas of sine,
\begin{equation}
   \begin{aligned}
     \label{eqt99}
     \sin(A+B) = \sin(A)\cos(B)+\cos(A)\sin(B),
   \end{aligned} 
\end{equation} 
and cosine, 
\begin{equation}
   \begin{aligned}
     \label{eqt99}
     \cos(A+B) = \cos(A)\cos(B)-\sin(A)\sin(B),
   \end{aligned} 
\end{equation}
results in:
\begin{equation}
   \begin{aligned}
     \label{eqt17}
     c_2^{V_i} = \frac{1}{\cos(\psi_m^G-\psi_i)}(x_{t,i}\sin(\psi_m^G) + (c_1^G-y_{t,i})\cos(\psi_m^G)).
   \end{aligned} 
\end{equation}

Note that if the straight line is perpendicular to the $x^{V_i}$ axis,
this equation will not have a solution. In other words, 
the $y^{V_i}$ intercept, denoted by $c_2^{V_i}$, cannot be determined. 

\subsubsection{Circles}
Let a circle be described by its center point $\pmb{C^G}(x_C^G, y_C^G)$ 
and has a radius of $R_C$ in the global coordinates. 
Since the radius does not change, only the center point 
is transformed using \eqref{eqt7}, resulting in:
\begin{equation}
   \begin{aligned}
     \label{eqt45}
     \pmb{C^{V_i}}
     =
     \begin{bmatrix}
      x_C^{V_i} \\
      y_C^{V_i} \\
     \end{bmatrix}
     =
     \begin{bmatrix}
      \cos(\psi_i) & \sin(\psi_i)  \\
      -\sin(\psi_i) & \cos(\psi_i) \\
     \end{bmatrix}
     \begin{bmatrix}
      x_C^G-x_{t,i} \\
      y_C^G-y_{t,i} \\
     \end{bmatrix}.
   \end{aligned} 
\end{equation}

\subsection{Pure Pursuit}
\label{sectionC}

\begin{figure}[!h]
   \flushleft
   \includegraphics[scale=0.23]{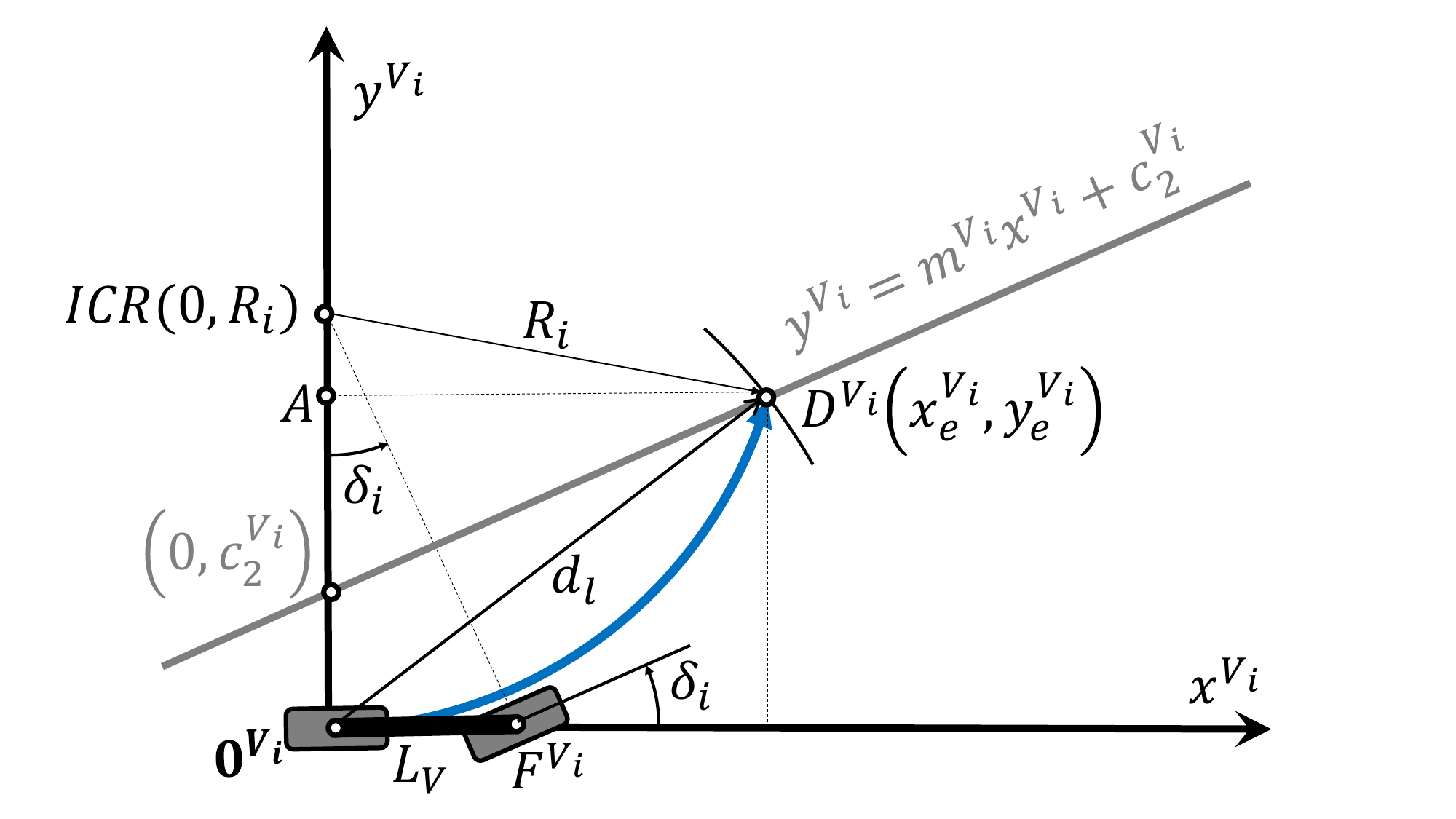}
   \caption{The look-ahead point $D^{V_i}$ in the vehicle's coordinates 
   also lies along the straight line and is located at a distance of $d_l$ from the 
   rear axle.}
   \label{fig3}
\end{figure}
The objective of this section is to determine the cross-track error $y_e^{V_i}$
and the corresponding steering angle $\delta_i$ for each sigma point 
$\pmb{\mathcal{X}_i}$. 
Employing the bicycle model within the pure pursuit algorithm, 
the vehicle is required to follow a circular path around the instantaneous center of rotation ($ICR$),
located at the point 
$(0,R_i)$ in the $i$-th vehicle's coordinates, as depicted in Figure \ref{fig3} \cite{Coulter92}.
At the end, it reaches the look-ahead point $D^{V_i}(x_e^{V_i}, y_e^{V_i})$, determined by 
the look-ahead distance $d_l$ from the rear axle, resulting in:
\begin{equation}
   \begin{aligned}
      \label{eqt18}
      d_l^2 = (x_e^{V_i})^2 + (y_e^{V_i})^2.
   \end{aligned} 
\end{equation}

By considering the triangle formed by points $A$, $D^{V_i}$, and ${ICR}$, 
we can establish the relationship: 

\begin{equation}
   \begin{aligned}
      \label{eqt21}
      R_i^2 = (x_e^{V_i})^2 + (R_i-y_e^{V_i})^2 \\ 
      = (x_e^{V_i})^2 + (y_e^{V_i})^2 - 2R_iy_e^{V_i} + R_i^2.
   \end{aligned} 
\end{equation}

Solving for $R_i$ yields:
\begin{equation}
   \begin{aligned}
      \label{eqt21}
      R_i = \frac{d_l^2}{2y_e^{V_i}}.
   \end{aligned} 
\end{equation}

Using the bicycle model, with the triangle formed by the points $ICR$, $0^{V_i}$, and $F^{V_i}$ 
as shown in Figure \ref{fig3},
we can derive the steering angle $\delta_i$ using the radius $R_i$, 
the wheelbase $L_V$ and \eqref{eqt21} as follows:
\begin{equation}
   \begin{aligned}
      \label{eqt23}
      \delta_i = \arctan\Biggl(\frac{L_V}{R_i}\Biggr) =  \arctan\Biggl(\frac{2y_e^{V_i}L_V}{d_l^2}\Biggr).
   \end{aligned} 
\end{equation}

Typically, the adaptive look-ahead distance $d_l$  can also be expressed as
proportional to the current vehicle speed $v_V$ 
with a proportionality constant $K_d$, given by:
\begin{equation}
   \begin{aligned}
      \label{eqt22}
      d_l = K_dv_V.
   \end{aligned} 
\end{equation}

From \eqref{eqt23}, we only have to calculate the cross-track error $y_e^{V_i}$ 
to find the steering angle $\delta_i$.
The determination of this error depends on the geometric type of road, 
and its derivation will be done separately as follows. 

\subsubsection{Straight Lines}

Since $(x_e^{V_i}, y_e^{V_i})$ lies on the straight line, 
as shown in Figure \ref{fig3}, it must satisfy the reformulated \eqref{eqt9}: 
\begin{equation}
   \begin{aligned}
      \label{eqt24}
      x_e^{V_i} = \frac{y_e^{V_i}-c_2^{V_i}}{m^{V_i}}.
   \end{aligned} 
\end{equation}
In the case where $m^{V_i} = 0$, we have the relation: 
\begin{equation}
   \begin{aligned}
      \label{eqt25}
       y_e^{V_i} = c_2^{V_i},
   \end{aligned} 
\end{equation}
otherwise, where $m^{V_i} \neq 0$ inserting \eqref{eqt24} in \eqref{eqt18}, 
solving the resulting quadratic equation and considering that 
$x_e^{V_i}$ is positive, we obtain the formula for $y_e^{V_i}$ as follows:
\begin{equation}
   \begin{aligned}
      \label{eqt26}
       y_e^{V_i} = \frac{c_2^{V_i} + m^{V_i} \sqrt{(1+(m^{V_i})^2)d_l^2-(c_2^{V_i})^2}}{1+(m^{V_i})^2}.
   \end{aligned} 
\end{equation}

\subsubsection{Circles}
\begin{figure}[!h]
   \flushleft
   \includegraphics[scale=0.23]{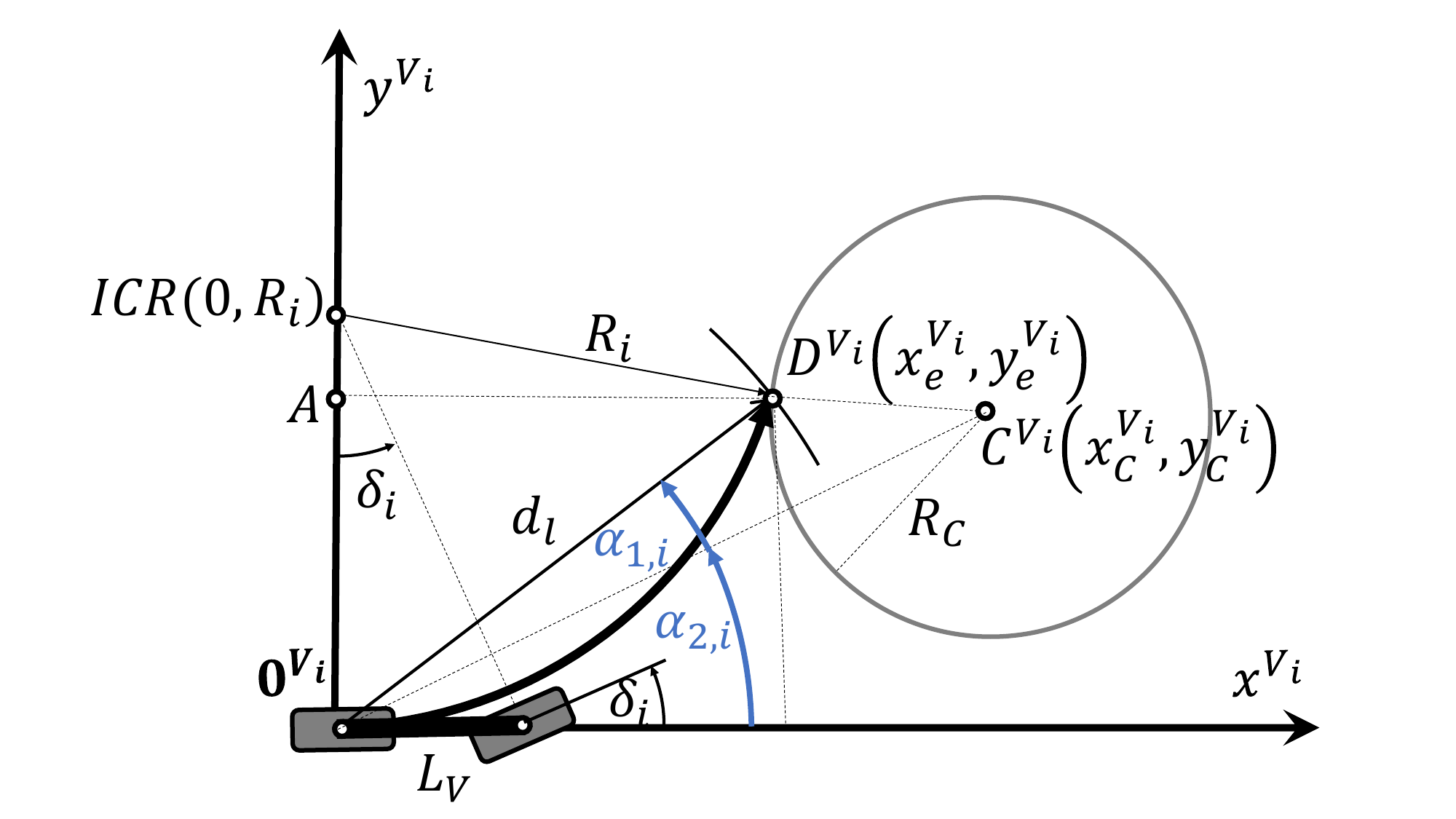}
   \caption{Pure pursuit with a look-ahead point, $D^{V_i}$,
   in the circle's circumference as a reference point for the vehicle.}
   \label{fig5}
\end{figure}

To calculate the cross-track error $y_e^{V_i}$,
we require two auxiliary angles $\alpha_{1,i}$, and $\alpha_{2,i}$, as can be seen in Figure \ref{fig5}.
By examining the triangle $0^{V_i}$, $D^{V_i}$, and ${C^{V_i}}$ 
and applying the law of cosines, the first angle $\alpha_{1,i}$ is obtained as follows:
\begin{equation}
   \begin{aligned}
      \label{eqt40}
      \alpha_{1,i} =  \arccos\Biggl(\frac{(x_C^{V_i})^2+(y_C^{V_i})^2+d_l^2-R_C^2}{2d_l\sqrt{(x_C^{V_i})^2+(y_C^{V_i})^2}}\Biggr).
   \end{aligned} 
\end{equation}
There are two solutions for $\alpha_{1,i}$ 
because $\cos(\alpha_{1,i}) = \cos(-\alpha_{1,i})$. The angle $\alpha_{2,i}$ is calculated as:
\begin{equation}
   \begin{aligned}
      \label{eqt41}
      \alpha_{2,i} =  \arctan\Biggl(\frac{y_C^{V_i}}{x_C^{V_i}}\Biggr).
   \end{aligned} 
\end{equation}

To maintain the vehicle's current direction as closely as possible, 
we need to use the smallest angle, determined by $\alpha_{min} \leftarrow \min|\alpha_{2,i} \pm \alpha_{1,i}|$.

As a result, this yields the cross-track error $y_e^{V_i}$: 
\begin{equation}
   \begin{aligned}
      \label{eqt42}
      y_e^{V_i} = d_l\sin(\alpha_{min}).
   \end{aligned} 
\end{equation}

\subsection{Calculate Weighted Steering Angle}
\label{sectionD}

Let $g_i$ represent the nonlinear transformation of the sigma point $\pmb{\mathcal{X}_i}$
to the steering angle $\delta_i= g_i(\pmb{\mathcal{X}_i})$, as depicted in Figure \ref{fig04} (b2).
The resulting weighted steering angle $\delta$ can then be obtained as follows:
\begin{equation}
   \begin{aligned}
      \label{eqt39}
      \mathcal{\delta} = \mathcal{W}_0\delta_0 + \mathcal{W}_{1...6}\sum_{i=1}^{6}\delta_i,
   \end{aligned} 
\end{equation}
where the weights $\mathcal{W}_i$ can be calculated according to \cite{Wan00}:
\begin{equation}
   \begin{aligned}
      \label{eqt38}
      \mathcal{W}_0 = \frac{\lambda_{UT}}{L_{UT}+\lambda_{UT}}, 
      \mathcal{W}_{1...6} = \frac{1}{2(L_{UT}+\lambda_{UT})}.
   \end{aligned} 
\end{equation}

\subsection{Update Pose}
\label{sectionE}

After obtaining the steering angle $\delta$, 
we rotate around the instantaneous center of rotation $ICR$ by the angle $\beta$, 
using the bicycle model, where $T$ denotes the update time:
\begin{equation}
   \begin{aligned}
      \label{eqt30}
      \beta = \frac{v_VT}{L_V}\tan(\delta), 
   \end{aligned} 
\end{equation}
which results in the translation: 
\begin{equation}
   \begin{aligned}
      \label{eqt31}
      \pmb{(t')}^{V_0}=
      \begin{bmatrix}
         (x')^{V_0} \\
         (y')^{V_0} \\  
        \end{bmatrix}
        =
        v_VT
        \begin{bmatrix}
         \cos(\beta)\\
         \sin(\beta) \\  
        \end{bmatrix},
   \end{aligned} 
\end{equation}
and in the orientation in the global coordinates:
\begin{equation}
   \begin{aligned}
      \label{eqt32}
      \psi' = \psi + \beta. 
   \end{aligned} 
\end{equation}

Using Equation \ref{eqt7} yields the global position:
\begin{equation}
   \begin{aligned}
      \label{eqt33}
      \pmb{(t')}^{G}
     =
     \begin{bmatrix}
      (x')^G \\
      (y')^G \\
     \end{bmatrix}
     =
     \begin{bmatrix}
      \cos(\psi) & -\sin(\psi) \\
      \sin(\psi) & \cos(\psi) \\  
     \end{bmatrix}
     \pmb{(t')}^{V_0}
     +
     \begin{bmatrix}
      x_t \\
      y_t \\
     \end{bmatrix}.
   \end{aligned} 
\end{equation}

Furthermore, 
the pose $\pmb{\mathcal{X}} = [\pmb{t}^T \ \psi]^T$ 
needs to be updated by drawing from the normal distribution ${\mathcal{N}}$  
with the covariance $\pmb{\Sigma_{V^*}}$ to account for uncertainties:
\begin{equation}
   \begin{aligned}
      \label{eqt46}
      \pmb{\mathcal{X}}
      \sim
      \mathcal{N} \Biggl(
         \begin{bmatrix}
            (x')^G \\
            (y')^G \\
            \psi'
           \end{bmatrix}
           ,
           \begin{bmatrix}
            \sigma_{x}^2 & 0 & 0 \\
            0 & \sigma_{y}^2 & 0 \\  
            0 & 0 & \sigma_{\psi}^2
           \end{bmatrix}
     \Biggr).
   \end{aligned} 
\end{equation}

After the pose update, the procedure outlined in Figure \ref{fig02} is repeated.

\section{SIMULATION}
Now, the test scenarios for both straight and circular
roads are defined, and the conventional pure pursuit
as well as the unscented transform pure pursuit algo-
rithms are qualitatively analyzed in the ideal case and
under uncertainty.

\subsection{Setup}

Implemented in Python, the straight road is defined 
as $y^G = 0$, with $m^G = 0$ and $c_1^G = 0$, 
thus lying completely in the $x^G$ coordinate. 
Otherwise, the circular road has a radius of $R_C = 5 \ m$ 
and a center point at $(x_C^G, y_C^G) = (0, R_C)$. 
In both cases, the vehicle is set to start at position
$\pmb{t} = [0 \ 0.5]^T \ m,$ oriented at $\psi = 0^{\circ}$. 

The parameters for the unscented transform are taken from \cite{Wan00}, 
with the dimensionality of the pose being $L_{UT} = 3$, $\alpha_{UT} = 0.001$, and $\kappa_{UT} = 0$. 
The look-ahead gain is set to $K_d = 1 \ s$, similar to \cite{Ohta16}. 
The vehicle's speed is $v_V = 1 \ m/s$, and the vehicle's length is $L_V = 1 \ m$.
The standard deviations for the summarized uncertainty 
are $\sigma_{x} = 0 \ m $, $\sigma_{y} = 0.1 \ m$, and $\sigma_{\psi} = 10^{\circ}$. 
Note that the lateral distance from the reference path is also constrained to 
model the filtering capabilities, such as those of Kalman filter families, in suppressing noise.
The update time is set to $T = 0.1 \ s$. 
Here, only a qualitative result analysis is conducted.

\subsection{Results}

\begin{figure}[!h]
   \centering
    {\epsfig{file = 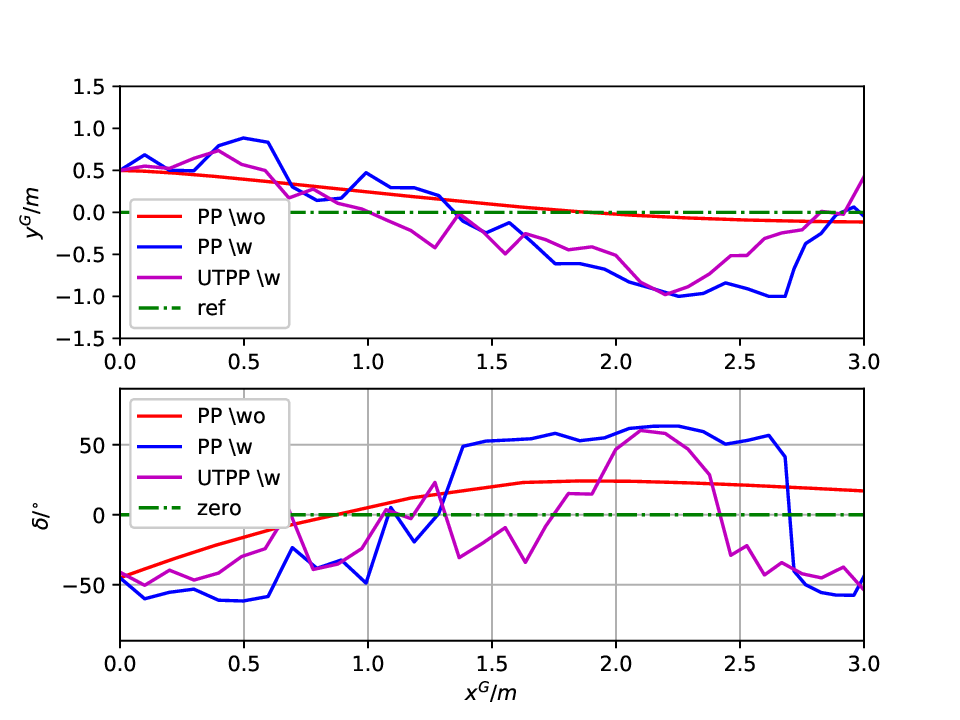, width = 8.25cm}}
   \caption{Comparing the paths $(x^G,y^G)$ and 
   the steering angle $\delta$ using the unscented 
   transform-based pure pursuit (UTPP with uncertainty) and 
   the pure pursuit (PP with or without uncertainty) 
   for straight roads ($\backslash wo$ 
   stands for "without noise", and $\backslash w$ for "with noise").}
   \label{fig9}
\end{figure}

\begin{figure}[!h]
   \centering
    {\epsfig{file = 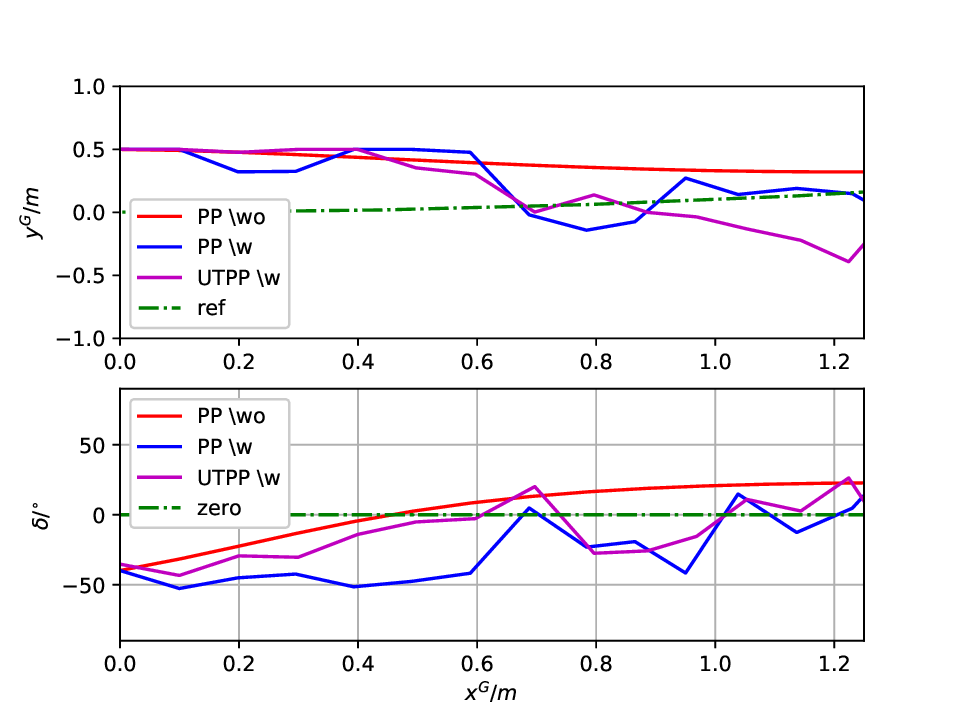, width = 8.25cm}}
    \caption{Comparing the results for circular roads, evaluating the paths $(x^G,y^G)$ 
    and the corresponding steering angles $\delta$ using the unscented 
    transform-based pure pursuit (UTPP) with uncertainty 
    and the conventional pure pursuit (PP) with and without uncertainty.}
   \label{fig10}
\end{figure}

In Figures \ref{fig9} and \ref{fig10}, the vehicle's positions $(x^G, y^G)$ over $x^G$ are illustrated. 
Under perfect sensing conditions (denoted as $wo$), represented by the red tracks, 
the vehicle effectively moves toward the reference path shown in green, 
converging with oscillations around it using the original pure pursuit (PP) algorithm. 
This behavior results from the steering angle $\delta$ calculated by the algorithm, 
where the vehicle initially steers right (indicated by a negative steering angle) 
to stay aligned with the predefined track, given its initial position is to the 
left of the path. The steering angle plot also oscillates around the zero angle, 
reflecting the vehicle's movement toward a straight trajectory.

When noise is introduced (denoted as $w$), both the conventional pure pursuit (PP) 
in blue and the unscented transform-based pure pursuit (UTPP) in magenta demonstrate similar tendencies. 
Initially, both algorithms steer the vehicle to the right to approach the reference path, 
as indicated by the negative steering angles. However, the path generated by the unscented 
transform-based pure pursuit appears to converge to the reference path more quickly than 
that of the conventional pure pursuit under uncertainty. 
This may be attributed to the use of multiple sigma points (or steering angles) in the UTPP, 
which aids the vehicle in returning to the track, as opposed to the conventional pure pursuit, 
which relies solely on a single calculated steering angle. 

However, as the vehicle approaches the track, 
certain sigma points may cause it to deviate from the path, 
as observed later in the simulation. 
Techniques such as varying the look-ahead distance could help mitigate this behavior, 
providing a potential avenue for investigation in future studies.

\section{CONCLUSIONS}
An unscented transform-based pure pursuit algorithm 
is proposed to address uncertainties in tracking reference 
paths across various road types. This method effectively 
assists in guiding the vehicle back to the track amidst noise. 
However, it may inadvertently steer the vehicle away from 
the track once the reference path is reached. This behavior 
stems from the algorithm's current lack of adaptability in determining the look-ahead distance, 
which is an area of active research.

Future improvements will focus on integrating dynamically changing look-ahead distances 
to enhance the algorithm's responsiveness. Additionally, efforts will be made to develop 
the algorithm to autonomously classify different road types. 
A generalized form of cross-track errors will be derived to accommodate 
various road geometries, including clothoid roads. Furthermore, 
the integration of the unscented Kalman filter for state observation 
will be explored for more practical applications. Finally, 
a quantitative analysis will be conducted to evaluate the performance and limitations of the proposed approach.

\bibliographystyle{apalike}
{\small
\bibliography{example}}

\end{document}